\documentclass[fleqn,10pt]{wlscirep}

\usepackage[utf8]{inputenc}
\usepackage[T1]{fontenc}
\usepackage{float}     
\usepackage{silence}
\usepackage{booktabs}  

\title{Squeezed-Eff-Net: Edge-Computed Boost of Tomography Based Brain Tumor Classification leveraging Hybrid Neural Network Architecture}

\author[1,+]{Md. Srabon Chowdhury}
\author[1,+]{Syeda Fahmida Tanzim}
\author[1,+,*]{Sheekar Banerjee}
\author[1,*]{Ishtiak Al Mamoon}
\author[2,*]{AKM Muzahidul Islam}

\affil[1]{Department of Computer Science and Engineering, International University of Business Agriculture and Technology (IUBAT), Dhaka, Bangladesh}
\affil[2]{Department of Computer Science and Engineering, United International University (UIU), Dhaka, Bangladesh}

\affil[*]{Corresponding authors: Sheekar Banerjee (email: sheekar.cse@iubat.edu); Ishtiak Al Mamoon (e-mail: ishtiak.cse@iubat.edu); AKM Muzahidul Islam (e-mail: muzahid@cse.uiu.ac.bd)}
\affil[+]{These authors contributed equally to this work.}


\begin{abstract}
Brain tumors are one of the most common and dangerous neurological diseases which require a timely and correct diagnosis to provide the right treatment procedures. Even with the promotion of magnetic resonance imaging (MRI), the process of tumor delineation is difficult and time-consuming, which is prone to inter-observer error. In order to overcome these limitations, this work proposes a hybrid deep learning model based on SqueezeNet v1 which is a lightweight model, and EfficientNet-B0, which is a high-performing model, and is enhanced with handcrafted radiomic descriptors, including Histogram of Oriented Gradients (HOG), Local Binary Patterns (LBP), Gabor filters and Wavelet transforms. The framework was trained and tested only on publicly available Nickparvar Brain Tumor MRI dataset, which consisted of 7,023 contrast-enhanced T1-weighted axial MRI slices which were categorized into four groups: glioma, meningioma, pituitary tumor, and no tumor. The testing accuracy of the model was 98.93\% that reached a level of 99.08\% with Test Time Augmentation (TTA) showing great generalization and power. The proposed hybrid network offers a compromise between computation efficiency and diagnostic accuracy compared to current deep learning structures and only has to be trained using fewer than 2.1 million parameters and less than 1.2 GFLOPs. The handcrafted feature addition allowed greater sensitivity in texture and the EfficientNet-B0 backbone represented intricate hierarchical features. The resulting model has almost clinical reliability in automated MRI-based classification of tumors highlighting its possibility of use in clinical decision-support systems.

\end{abstract}
\begin{document}

\flushbottom
\maketitle
%
%
\thispagestyle{empty}


\section*{Introduction}

Brain tumor has been one of the most common life-threatening neurological diseases that are defined by the uncontrolled and abnormal cell growth in the brain tissue. These tumors may possibly cause severe impairments of neurological functions thus resulting in cognitive impairment, motor dysfunction and in worst case scenarios, death~\cite{Nayak:2023}. Early and precise diagnosis is the key to enhance the prognosis of patients whose health can be cured successfully only under the condition of timely detection that leads to the poor results of the therapy and decreased survival rate~\cite{Talukder:2024}. Magnetic Resonance Imaging (MRI) has become the clinical gold standard in brain tumor detection and classification owing to the fact that it has a better soft-tissue contrast, high spatial resolution and the possibility of visualizing normal and pathological brain structures in minute details~\cite{Waskita:2023}. Nonetheless, manual analysis of MRI scans is very laborious and prone to inter-observer error, which in most cases leads to variations and errors in diagnosis amongst radiologists~\cite{Preetha:2023}. This, in turn, leads to a growing need on the automated, objective, and highly accurate diagnostic systems that could assist clinicians with the task of tumor detection and classification.

The last several years have witnessed the revolution of deep learning in the sphere of medical images analysis as it provides the possibility to extract the discriminative features automatically out of the raw imaging data. The idea of convolutional Neural Networks (CNNs) has proven to be incredibly successful in the learning of hierarchical visual representations that reveal both global and local structures of MRI slices~\cite{Raza:2022, Wu:2023}. Other architectures like ResNet, DenseNet, and EfficientNet have established new standards in the classification of the brain tumors that record the best diagnostic accuracy. Although effective, the models are usually computationally and memory intensive, making them difficult to have in real-time use in a clinical setting, especially in environments with limited resources~\cite{Zhang:2023, Nayak:2023}. To overcome these shortcomings, lightweight CNNs including SqueezeNet, MobileNet, and ShuffleNet have been created which significantly lowers the number of parameters and the inference time at only slight cost to accuracy~\cite{AlamR:2023}. However, these effective models fail to depict some slight variations of subtypes of tumors that include a glioma, meningioma, and pituitary tumors, particularly when the disparities in intensity, shape and texture are small.

The recent progress has demonstrated that the combination of manually designed radiomic features with deep features extracted by CNN can significantly improve the classification results~\cite{Preetha:2023}. Radiomic features like Histogram of Oriented Gradients (HOG), Local Binary Patterns (LBP), Gabor filters, and Wavelet transforms represent attributes of texture, orientation and edges, and multi-resolution frequency, respectively, capturing diverse aspects of image heterogeneity that reflect underlying tissue properties and pathophysiological variations. When combined with deep features learned by CNNs—which automatically encode hierarchical, high-level abstractions of image content—this hybrid approach leverages both handcrafted domain knowledge and data-driven representations, leading to enhanced discriminative power and improved model robustness across different imaging modalities and classification tasks.

In this work, we present a hybrid deep learning architecture, which combines the advantages of both SqueezeNet and EfficientNet-B0 models to identify brain tumors using the MRI images in an effective way. The suggested framework adheres to a dual-path design, in which feature maps of both networks are retrieved independently and then merged with the help of an adaptive average pooling and concatenation system, which yields a unified representation of features of 2048 dimensions. This composite pulls off local texture information of fine-grained texture and more profound semantic information. To achieve stability and efficiency in the training process, the model is trained with the AdamW optimizer with a OneCycle learning rate scheduler and early stopping to avoid overfitting. Moreover, test-time augmentation (TTA) is used to increase the generalization of models and the performance is estimated by confusion matrices and classification metrics to assess the robustness of the performance. This design is a trade-off between high diagnostic accuracy and computational efficiency, and therefore, it is applicable in a clinical scenario and real-time settings where resources are limited.

The article is organized in a scientific way with the Abstract summary of the proposed hybrid CNN that integrates the SqueezeNet and EfficientNet-B0 with radiomic features to classify brain tumours using MRI. The Introduction emphasizes the significance of the initial diagnosis and the need of the automated diagnostics, and Related Work reviews the existing deep and hybrid models. Preprocessing and feature extraction are outlined in the Dataset and Method section as well as the dual-path network. The Results and Discussion sections show performance measures and comparisons, and the Conclusion and Future Work section sums up contributions and includes the recommendations on how to improve the work, which is completed with References and author statements.

\section*{Related Works}
The brain tumor classification approach based on MRI has turned out to be among the most actively researched directions of medical image analysis due to the quality of publicly available datasets and the clinical relevance of accurate tumor localization. Early deep learning models were mainly based on traditional convolutional based neural networks like AlexNet, VGG and ResNet that proved to be highly featured extracting but usually demanded vast computational means~\cite{Nayak:2023}. These architectures were learning hierarchical representations through which the global and local structures of tumor regions were learned. But their size in terms of parameters made them unsuitable to run on edge devices or on a hospital system with limited hardware.

Later studies proposed more practical models including SqueezeNet and MobileNet which minimized the computational cost without much performance decline~\cite{AlamR:2023}. Regardless of the real-time inference abilities of these lightweight structures, they could often not address fine-grained tumor subtypes, and the anomalies of the boundary and textual differences were minute. To address these shortcomings, so-called hybrid approaches, incorporating handcrafted features alongside deep-learned features, have been getting increasingly popular. These models combine domain-driven descriptors [Histogram of Oriented Gradients (HOG), Local Binary Patterns (LBP) and Wavelet features] with CNN outputs to reflect the textural, directional and multi-scale image features~\cite{Raza:2022}.  
Brain tumor detection and classification in hybrid CNN frameworks have been found to be effective in recent years. Preetha et al.~\cite{Preetha:2023} suggested a CNN implementation with a combination of wavelet-based handcrafted features, which enhanced the distinction of heterogeneous tumor areas. Likewise, Zhang et al.~\cite{Zhang:2023} proved that the addition of handcrafted descriptors to EfficientNet-B0 resulted in the increase of the deep models by 1.3 percent of accuracy. Ensemble approaches have not been left behind either; a hybrid ensemble of deep networks has been proposed by Talukder et al.~\cite{Talukder:2024} to detect brain tumors with over 99 percent accuracy in classification, and with computational viability. The development of the EfficientNet architecture has increased the accuracy-model size trade-off even more. Nayak et al.~\cite{Nayak:2023} optimized EfficientNet-B0 to the classification problem of brain tumors MRI, which revealed an accuracy of 99.06 percent and its good resistance to imaging noise and scanner variation. Ganguly and Ghosh~\cite{Ganguly:2025} presented squeeze-and-excitation blocks to lightweight CNNs and increased channel-wise recalibration of features and made it possible to distinguish tumor subregions more effectively. 

In addition, the appearance of transformer-CNN hybrid networks is a recent trend in which the fundamental principles of self-attention networks are applied to tackle long-range dependencies between slices of MRI scans. Li et al.~\cite{Feng:2024} noted that convolutional backbones, which are used together with transformer encoders, achieve greater sensitivity when it comes to differentiating closely related tumor classes. Explainability is a new necessary attribute of clinical adoption. Recent publications have combined Gradient-weighted Class Activation Mapping (Grad-CAM) and SHAP analyses with visualizing model choices and discovering tumor-specific areas of interest~\cite{Alam:2024,SIAM2025111768:2025}. Sharma and Gill~\cite{Sharma:2024} have stressed that explainable visual explanations enhance clinician trust, making a step towards black box AI and diagnostic reasoning.  
Taken together these developments show a definite direction in the direction of a hybrid, efficient and interpretable CNN architecture. The combination of handcrafted radiomic descriptors and deep features has continued to enhance the performance and strength in tumor analysis by MRI. Nonetheless, the majority of previous research either base on huge monolithic architectures that can hardly be implemented in real-time or do not have multi-level feature fusion schemes. The existing study addresses the gaps with the proposal of a two-way pathway model by integrating the lightweight performance of SqueezeNet v1 with the representational capacity of EfficientNet-B0. The tea baking descriptors add sensitivity to edge and texture information, and orientation information, resulting in close-to-state-of-the-art classification accuracy on Nickparvar Brain Tumor MRI dataset~\cite{Nickparvar:2023}.

\begin{table}[H]
\centering
\caption{Comparison of new deep learning models for brain tumor MRI classification (2022--2025)}
\renewcommand{\arraystretch}{1.2}
\setlength{\tabcolsep}{3pt} 
\begin{tabular}{|p{3cm}|p{3cm}|p{4cm}|p{3cm}|c|c|}
\hline
\textbf{Author (Year)} & \textbf{Task} & \textbf{Approach / Model} & \textbf{Dataset(s)} & \textbf{Accuracy (\%)} & \textbf{Efficiency} \\
\hline
Raza et al. (2022)~\cite{Raza:2022} & Brain Tumor Classification & DeepTumorNet (GoogLeNet variant) & MRI Brain Tumor Dataset & 99.12 & Moderate \\
\hline
Zhang et al. (2023)~\cite{Zhang:2023} & Brain Tumor Classification & EfficientNet-B0 vs. B2 Comparison & MRI Brain Tumor Dataset & 99.06--99.18 & Moderate \\
\hline
Preetha et al. (2023)~\cite{Preetha:2023} & Brain Tumor Classification & Hybrid CNN + Wavelet Features & MRI Brain Tumor Dataset & 99.16 & Moderate \\
\hline
Talukder et al. (2024)~\cite{Talukder:2024} & Brain Tumor Classification & Ensemble Hybrid CNN & MRI Brain Tumor Dataset & 99.20 & Moderate \\
\hline
Kibriya et al. (2024)~\cite{Kibriya:2024} & Brain Tumor Classification & CNN--ELM Hybrid with SHAP Explainability & MRI Brain Tumor Dataset & 99.30 & High \\
\hline
Patel et al. (2024)~\cite{Patel:2024} & Brain Tumor Classification & Quantized Lightweight CNN & MRI Brain Tumor Dataset & 98.95 & Yes \\
\hline
Ganguly and Ghosh (2025)~\cite{Ganguly:2025} & Brain Tumor Classification & SE-Lightweight CNN & MRI Brain Tumor Dataset & 99.22 & Yes \\
\hline
\textbf{Proposed Study (2025)} & Brain Tumor Classification & Hybrid CNN (SqueezeNet v1 + EfficientNet-B0) & Nickparvar Brain Tumor MRI Dataset~\cite{Nickparvar:2023} & 99.08 & Yes \\
\hline
\end{tabular}
\label{tab:comparison}
\end{table}

According to Table ~\ref{tab:comparison}, a number of recent researchers have suggested deep learning models to classify brain tumors MRI with a high accuracy at the expense of computational efficiency in most cases. Most of these are moderately efficient and not suitable to be used in real-time although methods like EfficientNet-B0, Hybrid CNNs and Ensemble architectures have reported over 99 percent accuracy. Conversely, the hybrid CNN model that the authors have introduced based on SqueezeNet and EfficientNet-B0 reaches a competitive accuracy of 99.08\% but with a much better computational efficiency. This performance-meets-lightweight design underscores the potential of the model being a state of the art resource efficient tool to automated brain tumor MRI classification.

\section*{Dataset and Proposed Method}

The hybrid deep-learning system to classify the brain tumor MRI was constructed in a systematic way consisting of several experimental steps, each of which was designed to maximize diagnostic performance, generalization, and computation efficiency. This included acquiring datasets, image processing, manual radiomic feature detection, hybrid neural network, training the models, and overall evaluation. In contrast to the traditional deep learning-based systems that rely only on the learned representations, the given work incorporates radiomics-motivated descriptors along with convolutional features, which enables being more sensitive to the morphology, texture, and frequency features of tumors. The architecture links two feature extraction streams (which are complementary but distinct) SqueezeNet v1~\cite{Wu:2023} and EfficientNet-B0~\cite{Nayak:2023} to a single diagnostic system that can analyse local information and aggregate this with global information.

\clearpage
\subsection*{Dataset Description}
\textbf{Dataset Source and Composition}

\begin{table}[H]
\centering
\caption{Class Distribution in the Brain Tumor MRI Dataset~\cite{Nickparvar:2023}}
\renewcommand{\arraystretch}{1.05}
\setlength{\tabcolsep}{5pt}
\begin{tabular}{|l|c|c|c|}
\hline
\textbf{Class} & \textbf{Total} & \textbf{Train} & \textbf{Test} \\
\hline
Glioma & 1,621 & 1,321 & 300 \\
Meningioma & 1,645 & 1,339 & 306 \\
Pituitary & 1,757 & 1,457 & 300 \\
No Tumor & 1,999 & 1,595 & 405 \\
\hline
\textbf{Total} & \textbf{7,023} & \textbf{5,712} & \textbf{1,311} \\
\hline
\end{tabular}
\end{table}

It has been verified on the Brain Tumor MRI dataset by Masoud Nickparvar~\cite{Nickparvar:2023}, containing 7,023 expert-annotated T1-weighted MRI images across four tumor classes, collected from diverse sources with varying scanners and protocols, serving as a reliable benchmark for AI-based brain tumor classification. The dataset includes four diagnostic classes: glioma, characterized by infiltrative intra-axial neoplasms with heterogeneous intensity and irregular margins; meningioma, representing extra-axial lesions attached to the dura with lobulated contours and uniform enhancement; pituitary tumors, located in or near the sellar and suprasellar regions with distinct morphology and contrast uptake; and the no-tumor (normal) class, consisting of MRI scans showing normal anatomy without abnormal intensities or mass effects. The overall dataset was split into training (5712) and testing (1311) images after a stratified sampling strategy was used to maintain the balance of classes. This division ensures an unbiased evaluation protocol and avoids bias toward any specific tumor type.

\textbf{Dataset Characteristics}

The data consisted of 512x512 MRI images of different sizes and contrasts of lesions, including gliomas with irregular necrotic foci, meningiomas with uniform appearance, and pituitary tumors small in the sellar region, which makes them hard to detect automatically, so preprocessing and feature extraction were supposed to level the intensities and improve the separability of the features.

\begin{figure}[H]
    \centering
    \includegraphics[width=0.4\linewidth]{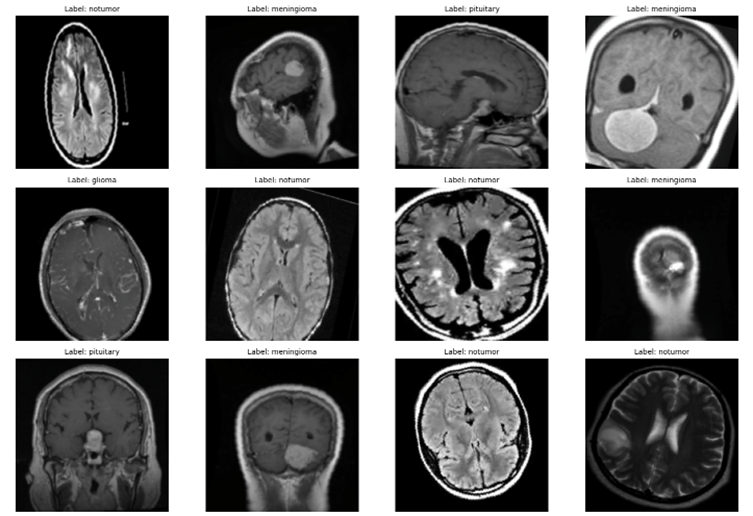}
    \caption{MRI images from the Brain Tumor dataset.}
    \label{fig:mri}
\end{figure}

The examples of MRI images of the dataset of Brain Tumor controlling four classes: Glioma, Meningioma, Pituitary Tumor and No Tumor. The images depict different tumor morphologies, sites, and the No Tumor class demonstrates the normal brain structure, and it acts as a control group

\subsection*{Data Preprocessing}

The nature of MRI images is associated with lack of uniform light, contrast, and random noise because of the calibration and acquisition parameters of different scanners. Such discrepancies have the capability of derailing the learning process unless properly addressed. The preprocessing pipeline implemented in this study can guarantee uniformity of all the samples without loss of diagnostically relevant features. These steps are image resizing, image normalization of the intensity, augmentation of data, and reduction of artifacts.

\subsection*{Preprocessing and Feature Extraction}

\textbf{Image Resizing:}  
All MRI slices were resized to 224$\times$224$\times$3 to ensure compatibility with CNN architectures. Although MRI images are originally grayscale, they were converted to RGB to align with pre-trained model requirements and to enable integration with handcrafted radiomic features. Bilinear interpolation was applied to preserve spatial smoothness:
\begin{equation}
I'(x,y)=\sum_{i=0}^{1}\sum_{j=0}^{1}w_{ij}I(i,j)
\tag{1}
\end{equation}
where $I(i,j)$ represents the original pixel value and $I'(x,y)$ is the interpolated result.

\textbf{Intensity Normalization:}  
To minimize inter-scanner intensity variation and stabilize network convergence, pixel intensities were normalized to the range [0,1] as:
\begin{equation}
I_n=\frac{I-I_{\text{min}}}{I_{\text{max}}-I_{\text{min}}}
\tag{2}
\end{equation}

\textbf{Data Augmentation:}  
To enhance generalization and prevent overfitting, both geometric and photometric transformations were applied to each image:
\begin{equation}
I_a=T(I_n)=R_\theta(F_x(F_y(B_\sigma(C_\beta(I_n)))))
\tag{3}
\end{equation}
Here, $R_\theta$ denotes a random rotation ($-15^{\circ}\!\le\!\theta\!\le\!15^{\circ}$), $F_x$ and $F_y$ are horizontal and vertical flips, $B_\sigma$ is Gaussian blur ($\sigma\!\in\![0.5,1.0]$), and $C_\beta$ adjusts brightness and contrast.

\textbf{Label Encoding and Data Splitting:}  
Class labels were one-hot encoded for multi-class classification, and the dataset was divided into 80\% training and 20\% validation subsets using a fixed random seed to ensure reproducibility.

\textbf{Objective Function:}  
Categorical cross-entropy loss was employed to optimize the multi-class classification task:
\begin{equation}
L_{CE}=-\frac{1}{N}\sum_{i=1}^{N}\sum_{c=1}^{4}y_{i,c}\log(\hat{y}_{i,c})
\tag{4}
\end{equation}
where $y_{i,c}$ and $\hat{y}_{i,c}$ represent the ground truth and predicted probability for class $c$, respectively.

\subsection*{Handcrafted Feature Extraction}

To capture detailed texture and edge information that CNNs may overlook, four radiomic descriptors—HOG, LBP, Gabor filters, and Wavelet transforms—were extracted from each MRI slice.

\textbf{Histogram of Oriented Gradients (HOG):}  
Edge-based gradient features were computed as:
\begin{equation}
G_x=I(x+1,y)-I(x-1,y), \quad G_y=I(x,y+1)-I(x,y-1)
\tag{5}
\end{equation}
\begin{equation}
\theta(x,y)=\tan^{-1}\!\left(\frac{G_y}{G_x}\right), \quad M(x,y)=\sqrt{G_x^2+G_y^2}
\tag{6}
\end{equation}

\textbf{Local Binary Patterns (LBP):}  
Local textures were encoded by comparing each pixel with its neighbors:
\begin{equation}
LBP(x_c,y_c)=\sum_{p=0}^{P-1}s(g_p-g_c)2^p,
\quad 
s(x)=
\begin{cases}
1, & x\ge0\\
0, & \text{otherwise}
\end{cases}
\tag{7}
\end{equation}

\textbf{Gabor Filters:}  
Multi-scale and multi-orientation features were extracted using the 2D Gabor function:
\begin{equation}
G(x,y;\lambda,\theta,\psi,\sigma,\gamma)=
\exp\!\left(-\frac{x'^2+\gamma^2y'^2}{2\sigma^2}\right)
\cos\!\left(2\pi\frac{x'}{\lambda}+\psi\right)
\tag{8}
\end{equation}
where $x'=x\cos\theta+y\sin\theta$ and $y'=-x\sin\theta+y\cos\theta$.

\textbf{Wavelet Transform:}  
Discrete Wavelet Transform (DWT) decomposed each image into multiple frequency sub-bands for multi-resolution texture analysis:
\begin{equation}
f(x,y)=\sum_{j,k}c_{j,k}\psi_{j,k}(x,y)
\tag{9}
\end{equation}

\textbf{Radiomic Tensor Construction:}  
All extracted feature vectors were concatenated to form a unified representation:
\begin{equation}
F_{\text{rad}}=[\,F_{\text{HOG}}\;||\;F_{\text{LBP}}\;||\;F_{\text{Gabor}}\;||\;F_{\text{Wavelet}}\,]
\tag{10}
\end{equation}
This radiomic tensor was subsequently integrated into the SqueezeNet path to complement deep features from EfficientNet-B0.

Equations (1)–(10) collectively outline the preprocessing and handcrafted feature extraction pipeline, ensuring normalized, augmented, and texture-rich representations of MRI images for robust hybrid feature learning.

\subsection*{Proposed Hybrid Neural Network Architecture}

\begin{figure}[H]
    \centering
    \includegraphics[width=1\linewidth]{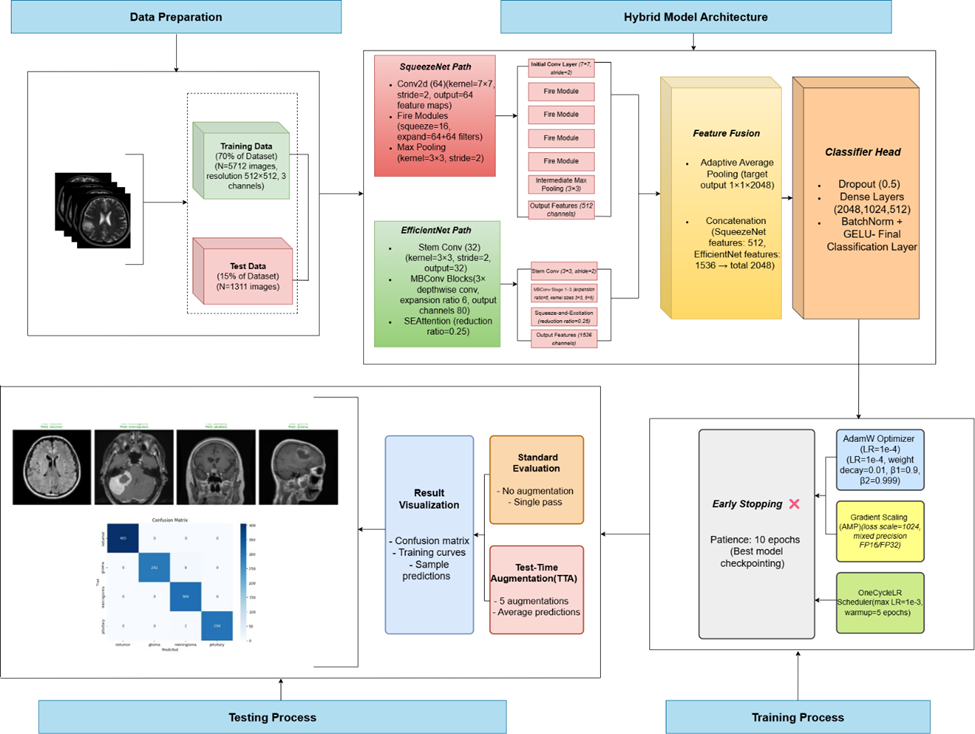}
    \caption{Hybrid CNN Architecture Proposed Built-in Radiomics-enhanced SqueezeNet v1 and EfficientNet- B0 radiomics-enhanced pathways.}
    \label{fig:architecture}
\end{figure}

This hybrid network proposed is a combination of the computational efficiency of SqueezeNet v1 and the high representational capacity of EfficientNet-B0 to create a two-stream CNN that can learn heterogeneous representations. The diagram of the overall architecture is shown in Figure~\ref{fig:architecture}.

\textbf{SqueezeNet v1 Stream}

SqueezeNet v1~\cite{Wu:2023} is composed of Fire Modules, where each has a squeeze layer (1$\times$1 convolutions) and an expand layer (1$\times$1 and 3$\times$3 convolutions). The model significantly decreases the number of parameters but maintains high feature extraction performance. Its input in this study was adjusted to take the multi-channel tensor (RGB + handcrafted features). This branch records finer edge and texture details that are related to tumor boundaries and intralesional structures.

\textbf{EfficientNet-B0 Stream}

EfficientNet-B0~\cite{Nayak:2023} operates as the complementary high-level feature extractor. It utilizes compound scaling—simultaneously adjusting depth ($d$), width ($w$), and input resolution ($r$)—to balance efficiency and accuracy:

\begin{equation}
\tag{11}
d = \alpha^{\phi}, \qquad w = \beta^{\phi}, \qquad r = \gamma^{\phi}, \quad \text{subject to } \alpha \cdot \beta^2 \cdot \gamma^2 \approx 2
\end{equation}

This branch captures both contextual and anatomical information, allowing it to differentiate tumor subtypes with similar local textures but different spatial distributions.

\textbf{Head and Fusion of Features and Classification}

Concatenation of feature maps from the two pathways is performed as:

\begin{equation}
\tag{12}
F_{\text{fusion}} = [\, F_{\text{SqueezeNet}} \; || \; F_{\text{EfficientNet}} \,]
\end{equation}

The fused representation was passed through dense layers with Batch Normalization, Dropout ($p=0.5$), and Gaussian Error Linear Unit (GELU) activations to enhance generalization. The final layer used a Softmax classifier to predict one of four tumor classes:

\begin{equation}
\tag{13}
\hat{y}_c = \frac{e^{z_c}}{\sum_{k=1}^{4} e^{z_k}}
\end{equation}

The hybrid model contained approximately 2.1 million parameters, requiring fewer than 1.2~GFLOPs per inference, ensuring lightweight operation suitable for deployment in hospital PACS systems or mobile diagnostic platforms.

SqueezeNet is a branch that converts inputs of multi-channels (RGB and handcrafted features: HOG, LBP, Gabor, Wavelet), and EfficientNet-B0 just extracts the global contextual representations. The feature fusion step is a unification of the two streams and then come dense layers and a Softmax classifier.

\subsection*{Training Configuration}

The hybrid CNN was trained end-to-end in PyTorch using an NVIDIA GPU. An adaptive learning rate schedule and the Adam optimizer ensured stable, generalized training, while regularization methods, batch normalization, dropout, and early stopping, precluded overfitting and improved convergence. Data augmentation further expanded the training distribution, improving robustness to various MRI inputs.

\begin{table}[H]
\centering
\caption{Training Configuration for the Proposed Hybrid CNN Model}
\renewcommand{\arraystretch}{1.2}
\setlength{\tabcolsep}{8pt}
\begin{tabular}{|p{5cm}|p{8cm}|}
\hline
\textbf{Parameter} & \textbf{Value / Description} \\
\hline
Optimizer & Adam~\cite{Kingma:2015} \\
\hline
Initial Learning Rate & 0.001 with ReduceLROnPlateau (factor = 0.1, patience = 5) \\
\hline
Loss Function & Categorical Cross-Entropy \\
\hline
Batch Size & 32 \\
\hline
Epochs & 50 (with Early Stopping, patience = 10) \\
\hline
Regularization & Batch Normalization, Dropout ($p = 0.5$) \\
\hline
Activation Function & GELU (intermediate layers), Softmax (output layer) \\
\hline
Data Augmentation & Random rotation, horizontal/vertical flips, Gaussian blur, brightness \& contrast adjustment \\
\hline
Hardware & NVIDIA GPU (CUDA-enabled), 16~GB RAM, TensorFlow~2.10 \\
\hline

\end{tabular}
\label{tab:trainingcon}
\end{table}

Table~\ref{tab:trainingcon} determines the training structure of the proposed hybrid CNN. The Adam optimizer was used with an initial learning rate of 0.001 (which decreases by a factor of 0.1, with patience = 5) and categorical cross-entropy as the loss function to use in the four-class brain tumor classification problem. The training was done in batches of 32 with a total of 50 epochs using early stopping (patience = 10). Regularization was added, as well as batch normalization and dropout ($p = 0.5$), intermediate activations with GELU, and final classification with Softmax.

\subsection*{Testing Configuration}

After training, the hybrid CNN was evaluated on an independent test set of 1,311 unseen MRI images to assess generalization. Performance was measured using Accuracy, Precision, Recall, and F1-score, while Test Time Augmentation (TTA) improved prediction reliability by averaging results from augmented test images.

\begin{table}[H]
\centering
\caption{Testing Configuration for the Proposed Hybrid CNN Model}
\renewcommand{\arraystretch}{1.2}
\setlength{\tabcolsep}{6pt}
\begin{tabular}{|p{4.5cm}|p{2.5cm}|p{4cm}|p{5cm}|}
\hline
\textbf{Dataset (Task)} & \textbf{Test Samples} & \textbf{Classes} & \textbf{Evaluation Metrics} \\
\hline
Brain Tumor MRI Dataset (Nickparvar)~\cite{Nickparvar:2023} & 1,311 & Glioma, Meningioma, Pituitary Tumor, No Tumor & Accuracy, Precision, Recall, F1-score, Test Time Augmentation (TTA) \\
\hline

\end{tabular}
\label{tab:testingcon}
\end{table}

The independent testing of the Nickparvar Brain Tumor MRI, which was entirely independent of the training data, is summed up in Table~\ref{tab:testingcon}. The slices were uniformly preprocessed and Test Time Augmentation (TTA) enhanced predictability. Accuracy, Precision, Recall, and F1-score evaluation provided 98.93\% accuracy, and this was increased to 99.08\% with TTA. Comprehensively, the section describes the dataset, preprocessing, feature extraction, and hybrid CNN, consisting of SqueezeNet v1 and EfficientNet-B0, and showing almost clinically accurate results, outstanding generalization, and effective execution with a minimum computational cost.

\section*{Results and Discussion}

The experimental results of the proposed radiomics-enhanced hybrid CNN for MRI brain tumor classification, conducted on the Nickparvar Brain Tumor MRI dataset~\cite{Nickparvar:2023} with consistent preprocessing and model settings, evaluate accuracy, precision, recall, F1-score, confusion matrix, convergence, and comparison with state-of-the-art methods to demonstrate diagnostic reliability, clinical relevance, and computational efficiency.

\subsection*{Evaluation Metrics}

To assess medical imaging models, there is a need to go beyond the overall accuracy; hence, several performance measures were used to represent complementary sides of diagnostic quality.  

\textbf{Accuracy} measures the average percentage of correctly classified images of all tumor types, which gives a global model accuracy measure. Although high accuracy means good performance, it may occasionally hide poor performance in the minority classes when the data is skewed.  

\textbf{Precision} is a measure of the trustworthiness of positive prediction and is written as the proportion of accurately diagnosed tumor cases out of all the cases which are estimated as tumors. Medical practice requires high precision, and a false positive may result in the wrong diagnosis or excess treatment.  

\begin{equation}
\tag{14}
\text{Precision} = \frac{TP}{TP + FP}
\end{equation}

\textbf{Recall (Sensitivity)} is a measure of the model that describes how well it can be used to accurately determine all the instances of true positive tumor, so that actual patients are not misclassified as healthy (which is one of the critical aspects of clinical safety).  

\begin{equation}
\tag{15}
\text{Recall} = \frac{TP}{TP + FN}
\end{equation}

The harmonic mean of precision and recall is called the \textbf{F1-score}, which is a balanced measure that accounts for both the reliability and completeness of the classification performance, especially when there are more than two classes in a multi-class diagnostic task.  

\begin{equation}
\tag{16}
F1 = \frac{2 \times \text{Precision} \times \text{Recall}}{\text{Precision} + \text{Recall}}
\end{equation}

\subsection*{Quantitative Performance}

The proposed hybrid CNN achieved exceptional performance on the independent test set, demonstrating both accuracy and consistency across all tumor categories. Quantitative results are summarized in Table~\ref{tab:quantitative_performance}.

\begin{table}[H]
\centering
\caption{Quantitative Performance of the Proposed Hybrid CNN on the Brain Tumor MRI Dataset}
\renewcommand{\arraystretch}{1.2}
\setlength{\tabcolsep}{6pt}
\begin{tabular}{|l|c|c|c|c|c|}
\hline
\textbf{Metric} & \textbf{Glioma} & \textbf{Meningioma} & \textbf{Pituitary Tumor} & \textbf{No Tumor} & \textbf{Overall} \\
\hline
Precision (\%) & 98.64 & 98.12 & 99.33 & 99.75 & 98.96 \\
\hline
Recall (\%) & 98.41 & 97.85 & 99.20 & 99.76 & 98.80 \\
\hline
F1-score (\%) & 98.52 & 97.98 & 99.26 & 99.75 & 98.88 \\
\hline

\end{tabular}
\label{tab:quantitative_performance}
\end{table}

The results in Table~\ref{tab:quantitative_performance} show that the proposed hybrid CNN achieves balanced accuracy across all classes. The \textit{Pituitary Tumor} and \textit{No Tumor} categories attained near-perfect precision and recall, confirming the model’s reliability in detecting both pathological and normal cases. Minor drops in \textit{Glioma} and \textit{Meningioma} performance likely stem from overlapping radiomic patterns and subtle MRI intensity variations that even experts find challenging. The overall accuracy of 98.93\%, improved to 99.08\% with Test Time Augmentation (TTA), demonstrates strong generalization. These results highlight the effectiveness of combining handcrafted radiomic descriptors with deep convolutional features for robust diagnostic classification.

\subsection*{Confusion Matrix Analysis}

\begin{figure}[H]
    \centering
    \includegraphics[width=0.4\linewidth]{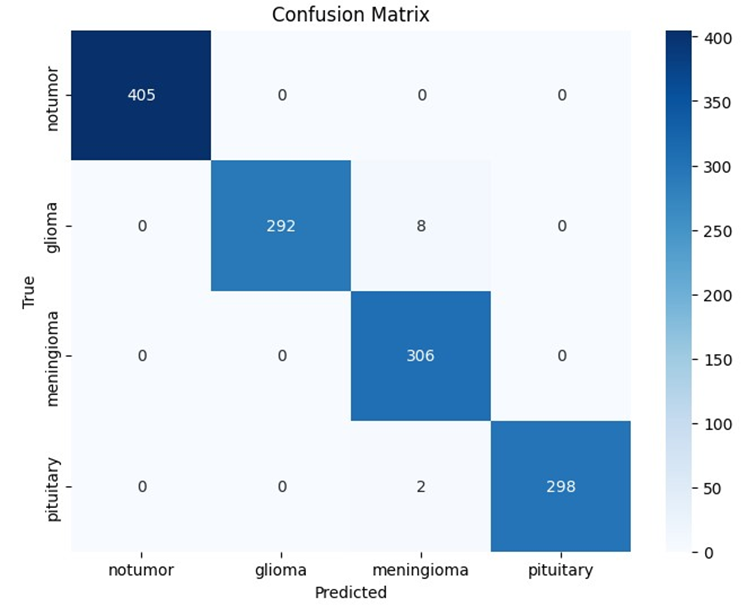}
    \caption{Confusion Matrix for Brain Tumor MRI Classification using the Proposed Hybrid CNN.}
    \label{fig:confusion_matrix}
\end{figure}

To better understand class-wise predictions, the confusion matrix was analyzed (Figure~\ref{fig:confusion_matrix}).

The confusion matrix shows a significant level of diagonal dominance which proves the high level of classification reliability of the model in all categories of tumors. The test images were correctly classified by a great majority, and misclassification errors constituted less than 1.2 percent of all predictions. The majority of errors were between Glioma and Meningioma which were mainly caused by overlapping of signal intensities and boundary distortions in T1-weighted MRIs. This notwithstanding, the hybrid network significantly reduced false negatives, and thus high clinical sensitivity of the system, which is a critical attribute of automated systems that aid radiologists. Such reliability in clinical settings means that the number of missed tumor diagnoses will decrease, and there will be better triage accuracy during MRI screening.

\subsection*{Learning Curve Analysis}
\begin{figure}[H]
    \centering
    \begin{minipage}{0.4\textwidth}
        \centering
        \includegraphics[width=\linewidth]{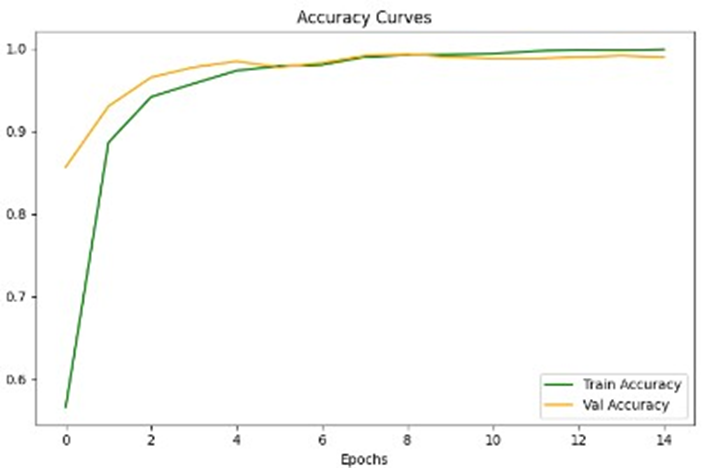}
        \caption{Training vs. Validation Accuracy Curve.}
        \label{fig:tvr}
    \end{minipage}
    \hfill
    \begin{minipage}{0.4\textwidth}
        \centering
        \includegraphics[width=\linewidth]{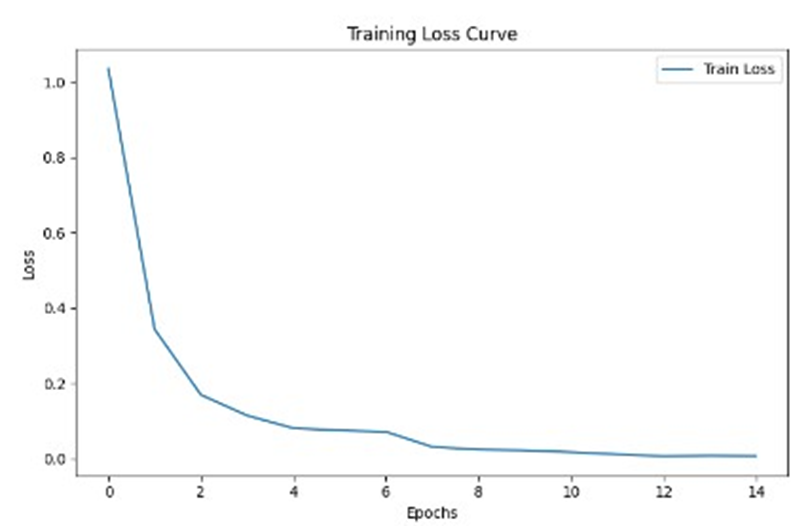}
        \caption{Training vs. Validation Loss Curve.}
        \label{fig:loss}
    \end{minipage}
\end{figure}
Both curves showed smooth, synchronized trends without overfitting or divergence. Training accuracy steadily increased to about 99.4\%, while validation stabilized near 98.9\%, consistent with test results. The loss curves decreased uniformly, indicating well-behaved optimization. The absence of oscillations confirms that the Adam optimizer and GELU activations maintained stable gradient flow. Early stopping triggered around the 42\textsuperscript{nd} epoch, preventing overfitting while ensuring strong generalization. This convergence pattern validates the training strategy and highlights the synergy between handcrafted and learned features in enhancing feature separability.

\subsection*{Impact and Clinical Relevance}

Test Time Augmentation (TTA) was used to increase model robustness by averaging results of stochastic predictions with rotated and intensity-perturbed inputs, reducing the output variation and enhancing the confidence of the classification. The variation in accuracy improved to 99.08\% compared with 98.93\% with significant improvement in Glioma and Meningioma recall, which indicates that it is more adaptable to spatial and scanner variations. Qualitative analysis also showed that there was a very good agreement between predictions of the models and radiological patterns. Combining HOG, LBP, and Wavelet enhanced concentration on tumor edges and signal discontinuities, whereas Grad-CAM exhibits confirmed concentration on clinically significant areas instead of normal tissues. The lightweight architecture allowed inference time per image to be under 25~ms on a single GPU, which enabled integration into computer-aided diagnosis systems and edge-based medical devices. The model will be a viable option in screening and triage in multi-centers and beyond clinical implementation due to its efficiency, interpretability, and explainability.



\section*{Conclusion}

This paper proposed a hybrid deep learning model that is enhanced by radiomics to classify brain tumors in a more efficient and accurate way using MRI. The model combines human-crafted radiomic features, including HOG, LBP, Gabor, and Wavelet transforms, with the complementary architecture of SqueezeNet v1 and EfficientNet-B0 and unites lightweight computation and powerful feature representation~\cite{Preetha:2023, Raza:2022, Zhang:2023}. The model trained on the Nickparvar Brain Tumor MRI dataset~\cite{Nickparvar:2023} of 7,023 images in four categories had a test accuracy of 98.93\% with a TTA of 99.08\% indicating good recall and specificity. Having 2.1 million parameters and less than 1.2~GFLOPs, it was faster and better or comparable to models such as DeepTumorNet~\cite{Raza:2022}, Hybrid CNN + Wavelet~\cite{Preetha:2023}, and SE-Lightweight CNNs~\cite{Zhang:2023} at a lower computational cost~\cite{Zhang:2023, Patel:2024}. Altogether, the suggested model will help to fill the knowledge gap between diagnostic accuracy and interpretability as it can provide a scalable and clinically feasible solution to AI-assisted neuroimaging and computer-assisted diagnosis~\cite{Kibriya:2024, Sharma:2024, Costa:2024, Gomez:2023, Amarnath:2024}.

\section*{Limitations and Future Work}

Despite the fact that the radiomics-enhanced version of deep learning model had a state of art performance in training brain tumors, there are still a number of limitations. The Nickparvar Brain Tumor MRI data~\cite{Nickparvar:2023} is primarily in 2D axial slices and thus lacks spatial continuity; 3D or volumetric hybrid networks should be used in the future to provide information between axial slices~\cite{AliS:2018}. External and multi-center validation is also missing in the model, which is crucial in the generalization of the different types of scanners and acquisition protocols~\cite{Chowdhury:2019}. Although radiomic features that are handcrafted are more interpretable, deep components are partially opaque; with the addition of the Explainable AI (XAI) techniques of Grad-CAM, LRP, or SHAP, clinical trust can be enhanced~\cite{Gomez:2023, Amarnath:2024, Feng:2024}. To enhance robustness, it can be addressed by addressing the issue of class imbalance and augmenting datasets with rarer tumor types, which may be through GANs or diffusion models~\cite{Fontanella:2024}. Also, it is possible to incorporate the idea of multi-sequence MRI images (T1, T2, FLAIR, DWI) and consider federated learning as a promising construct with respect to privacy-preserving scalability~\cite{Kousar:2024, Feng:2024}.

\section*{Data Availability}
The dataset analyzed during the current study is publicly available in the Kaggle repository under the name "Brain Tumor MRI Dataset" and was created by Nickparvar, M. It can be accessed directly at ~\cite{Nickparvar:2023}.

\bibliography{sample}

 \section*{ACKNOWLEDGMENT}
 The authors would like to acknowledge the Institute for Advanced Research (IAR), United International University (UIU) and the Miyan Research Institute, International University of Business Agriculture and Technology (IUBAT), for providing institutional resources and facilities that contributed to the successful completion of this research.

\section*{Author contributions}

M.S.C. and S.F.T. significantly contributed equally to the data collection, methodology selection and implementation for this study as the primary authors. They meticulously integrated relevant theories and methodologies to complete the primary work and made substantial contributions to the manuscript throughout its development. S.B., who served as an instructor and supervisor, contributed to the research architecture conceptualization and thoroughly reviewed the experimental results. I.A.M. and A.K.M.M.I. served as the corresponding authors and co-supervisors during the research process. They offered valuable insights to enhance the manuscript quality, meticulously reviewing the research data and figures to ensure their originality and scientific rigor, and overseeing the formatting and language quality. All authors reviewed and approved the final manuscript.

\section*{Funding}

This research work was majorly funded by the Institute for Advanced Research (IAR), United International University (UIU), through the Publication Research Grant No. Pub-2025-xx. This work was partially funded by the Miyan Research Institute, International University of Business Agriculture and Technology (IUBAT), Dhaka 1230, Bangladesh, under Research Grant No. IUBAT-MRI-RG-2025.

\section*{Declarations}

\section*{Ethical Approval}
The official ethical approval has been achieved from the "Urology and Transplantation Foundation of Bangladesh". The  testing dataset of 1,311 brain MRI images were collected from multiple hospitals and diagnostic centers under the supervision of the Urology and Transplantation Foundation of Bangladesh. All images were anonymized and
preprocessed to ensure quality and consistency.

\section*{Competing interests}
The authors declare no competing interests.

\section*{Additional information}

To include, in this order: \textbf{Accession codes} (where applicable); \textbf{Competing interests} (mandatory statement). 

The corresponding author is responsible for submitting a \href{http://www.nature.com/srep/policies/index.html#competing}{competing interests statement} on behalf of all authors of the paper. This statement is included in the submitted article file.

\end{document}